\documentclass[10pt,twocolumn,letterpaper]{article}

\usepackage{iccv}              

\iffalse
\newcommand{\dimpp}[1]{\textcolor{blue}{[DP: #1]}}
\newcommand{\onur}[1]{\textcolor{magenta}{[OK: #1]}}
\newcommand{\kaixuan}[1]{\textcolor{green}{[KL: #1]}}
\else
\newcommand{\dimpp}[1]{}
\newcommand{\onur}[1]{}
\newcommand{\kaixuan}[1]{}
\fi

\usepackage{algpseudocode}
\usepackage{mathtools}
\usepackage{float}
\usepackage{dsfont}
\usepackage{amsmath,amssymb}
\usepackage{multirow}
\usepackage{graphicx}

\makeatletter
\renewcommand\paragraph[1]{%
\vspace{0mm}\noindent\textbf{#1}%
}
\makeatother

\definecolor{iccvblue}{rgb}{0.21,0.49,0.74}
\usepackage[pagebackref,breaklinks,colorlinks,allcolors=iccvblue]{hyperref}

\def\paperID{25} 
\def\confName{ICCV}
\def\confYear{2025}

\title{AutoQ-VIS: Improving Unsupervised Video Instance Segmentation via Automatic Quality Assessment}

\author{
Kaixuan Lu$^{1}$ \quad Mehmet Onurcan Kaya$^{1,2}$ Dim~P.~Papadopoulos$^{1,2}$\\
$^{1}$Technical University of Denmark \quad 
$^{2}$Pioneer Centre for AI\\
\texttt{\small s232248@student.dtu.dk, monka@dtu.dk, dimp@dtu.dk}
}

\begin{document}

\maketitle
\begin{abstract}
Video Instance Segmentation (VIS) faces significant annotation challenges due to its dual requirements of pixel-level masks and temporal consistency labels. While recent unsupervised methods like VideoCutLER eliminate optical flow dependencies through synthetic data, they remain constrained by the synthetic-to-real domain gap. We present AutoQ-VIS, a novel unsupervised framework that bridges this gap through quality-guided self-training. Our approach establishes a closed-loop system between pseudo-label generation and automatic quality assessment, enabling progressive adaptation from synthetic to real videos. Experiments demonstrate state-of-the-art performance with 52.6 $\text{AP}_{50}$ on YouTubeVIS-2019 \texttt{val} set, surpassing the previous state-of-the-art VideoCutLER by 4.4$\%$, while requiring no human annotations. This demonstrates the viability of quality-aware self-training for unsupervised VIS.
%
The source code of our method is available at \href{https://github.com/wcbup/AutoQ-VIS}{here}.

\end{abstract}    
\section{Introduction}
\label{sec:intro}

\begin{figure}[t]
    \centering
    \includegraphics[width=1\linewidth]{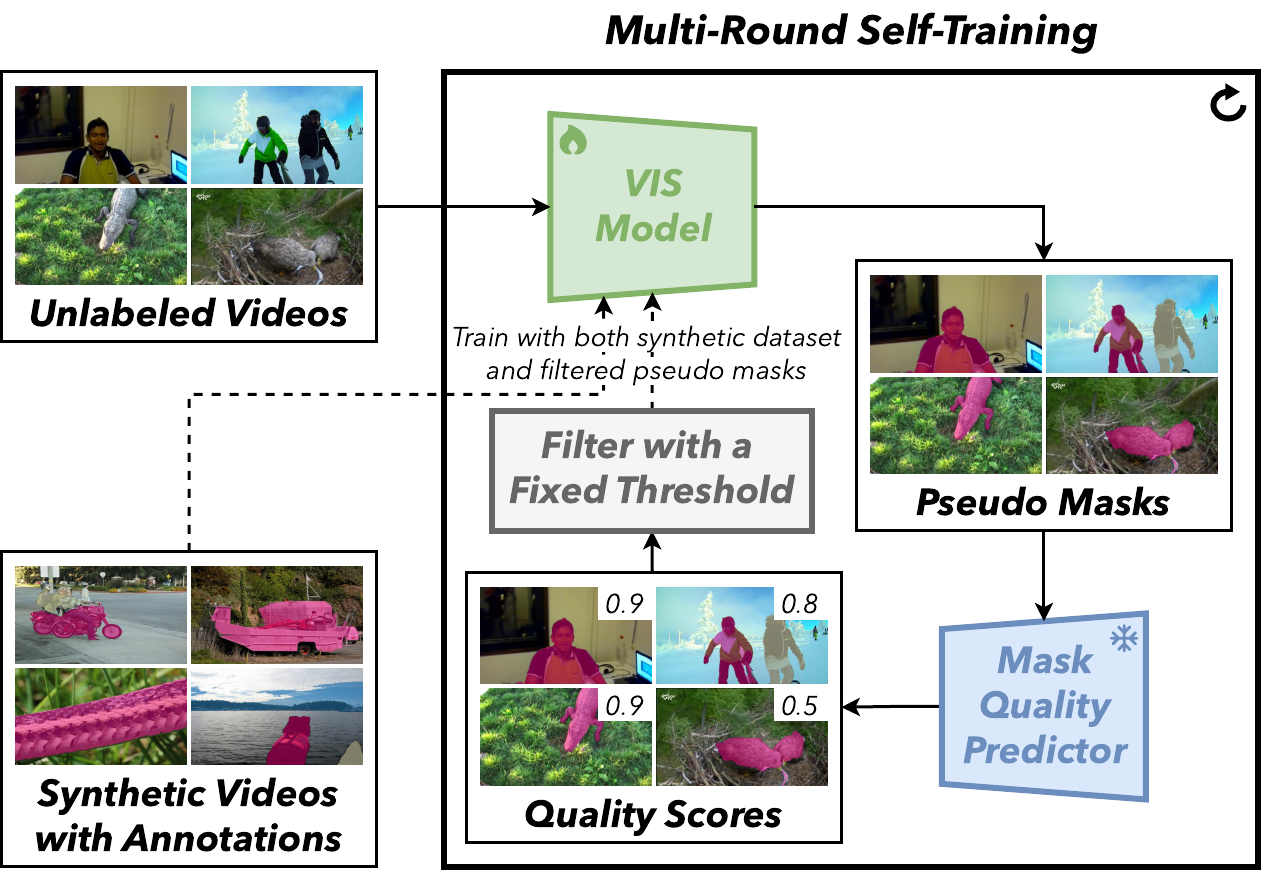}
    \caption{
        \textbf{AutoQ-VIS overview.} In the initial training stage, both the VIS model and the mask quality predictor are trained on synthetic videos with pseudo  annotations~\cite{videocutler}. During the multi-round self-training stage, the VIS model generates pseudo masks on unlabeled videos, which are then scored by the frozen quality predictor. Pseudo masks with high predicted quality are selected and added to the training set. The VIS model is subsequently retrained on both the synthetic data and the selected pseudo labels, enabling iterative refinement and progressive performance gains.
    }
    \label{fig:overall}
\end{figure}

Video Instance Segmentation (VIS) is the challenging task of simultaneously detecting, segmenting, and tracking object instances across video sequences~\cite{ytvis2019,wu2022seqformer,lin2020video,heo2022vita,wang2021end,zhang2023dvis}. This capability is fundamental for scene understanding in applications ranging from autonomous driving~\cite{driving_vis_dataset} to video content editing~\cite{vis_survey}. However, training high-performance VIS models typically requires pixel-level annotations across all frames~\cite{ytvis2019}. This process is expensive due to the labor-intensive nature of annotating temporal consistency and instance identities. As a result, there is an urgent need to develop unsupervised video instance segmentation frameworks that can accurately interpret video content and function effectively across diverse, unlabeled environments.

Prior work~\cite{motiongroup, emdriven, guessmove, bootstrapping, highly_probable_positive_features, transport_network, dense_video_seg} in unsupervised video segmentation predominantly addresses Video Object Segmentation (VOS), focusing on separating a single foreground object via motion or consistency cues. While OCLR~\cite{oclr} introduces unsupervised VIS that supports multiple instances, its predefined object count during training prevents dynamic adaptation to varying instances during inference. Furthermore, prior approaches~\cite{motiongroup, emdriven, guessmove, bootstrapping, oclr} rely on optical flow estimators (e.g.,  RAFT~\cite{raft}) that are trained on human-annotated datasets. VideoCutLER~\cite{videocutler} marks a breakthrough in unsupervised VIS and achieves unprecedented performance by demonstrating multi-instance segmentation without optical flow dependencies. Its core innovation lies in synthetic video generation via spatial augmentations of CutLER~\cite{cutler} pseudo-labels from ImageNet~\cite{imagenet}. However, VideoCutLER remains constrained by synthetic-to-real domain gaps and static instance modeling, i.e., the synthetic videos lack natural and realistic motion patterns. 

Building upon VideoCutLER's synthetic data paradigm, which generates training videos through spatial augmentations of static image pseudo-labels, we introduce AutoQ-VIS to address its critical domain gap limitation. While VideoCutLER's synthetic videos provide initial instance awareness, they lack natural motion patterns and real-world appearance variations, hindering adaptation to authentic video dynamics. Our framework bridges this synthetic-to-real domain gap through a self-training loop that progressively adds quality-filtered pseudo-labels from unlabeled real videos. Inspired by Mask Scoring R-CNN~\cite{maskscore}, which directly predicts mask quality scores via an auxiliary branch, we implement a quality assessment module for pseudo-label filtering of instance masks.
AutoQ-VIS advances unsupervised video instance segmentation through an iterative self-training paradigm with quality-aware pseudo-label selection (\cref{fig:overall}). The system initializes using synthetic video data from VideoCutLER, which provides pseudo-labels to bootstrap a VideoMask2Former~\cite{mask2former, videomask2former} VIS model and a specialized mask quality predictor (\cref{sec:initial_model}). The mask quality predictor 
estimates mask IoU quality scores by analyzing frame-level features and mask predictions. During multi-round optimization, the VIS model generates pseudo-labels on unlabeled videos, which are then scored by the quality predictor. High-quality pseudo-labels surpassing a fixed threshold are progressively incorporated into the training set, enabling dataset augmentation without any human supervision (\cref{sec:multi_round_training}). To enhance the mask head training, we employ a DropLoss that zeros out mask losses whose maximum ground‑truth overlap falls below 0.01 (\cref{sec:droploss}).  By alternating rounds of VIS training (with occasional weight resets) and quality‑based dataset expansion, AutoQ‑VIS dynamically enriches its training dataset and steadily sharpens segmentation performance.

Our key contributions are threefold: \textbf{(1) Annotation-Free VIS Framework:} We propose AutoQ-VIS, an unsupervised framework that overcomes annotation dependency through cyclic pseudo-label refinement with automated quality control, enabling video instance segmentation training directly from unlabeled videos. \textbf{(2) Automatic Quality Assessment:} We propose a simple quality predictor that reliably filters pseudo labels across self-training rounds. \textbf{(3) New State-of-the-art Performance:} Our AutoQ-VIS archives 52.6 ${\text{AP}_{50}}$ on YouTubeVIS-2019~\cite{ytvis2019} \texttt{val} split, surpassing the previous state-of-the-art VideoCutLER~\cite{videocutler} by 4.4 ${\text{AP}_{50}}$.

\section{Method}


AutoQ-VIS operates through three stages: (1) \textbf{Initial Training} (\cref{sec:initial_model}): Jointly pretrain VideoMask2Former and the mask quality predictor on VideoCutLER's synthetic videos; (2) \textbf{Multi-Round Self-Training} (\cref{sec:multi_round_training}): Iteratively generate pseudo-labels on real unlabeled videos, filter via quality scores, and augment training data; (3) \textbf{DropLoss} (\cref{sec:droploss}): Suppress low-IoU mask predictions to enhance mask head training. This quality-guided pipeline progressively improves segmentation accuracy without any human annotations.

\dimpp{we need an initial paragraph here as an overview that explains the pipeline!}

\begin{figure}[t]
    \centering
    \includegraphics[width=1\linewidth]{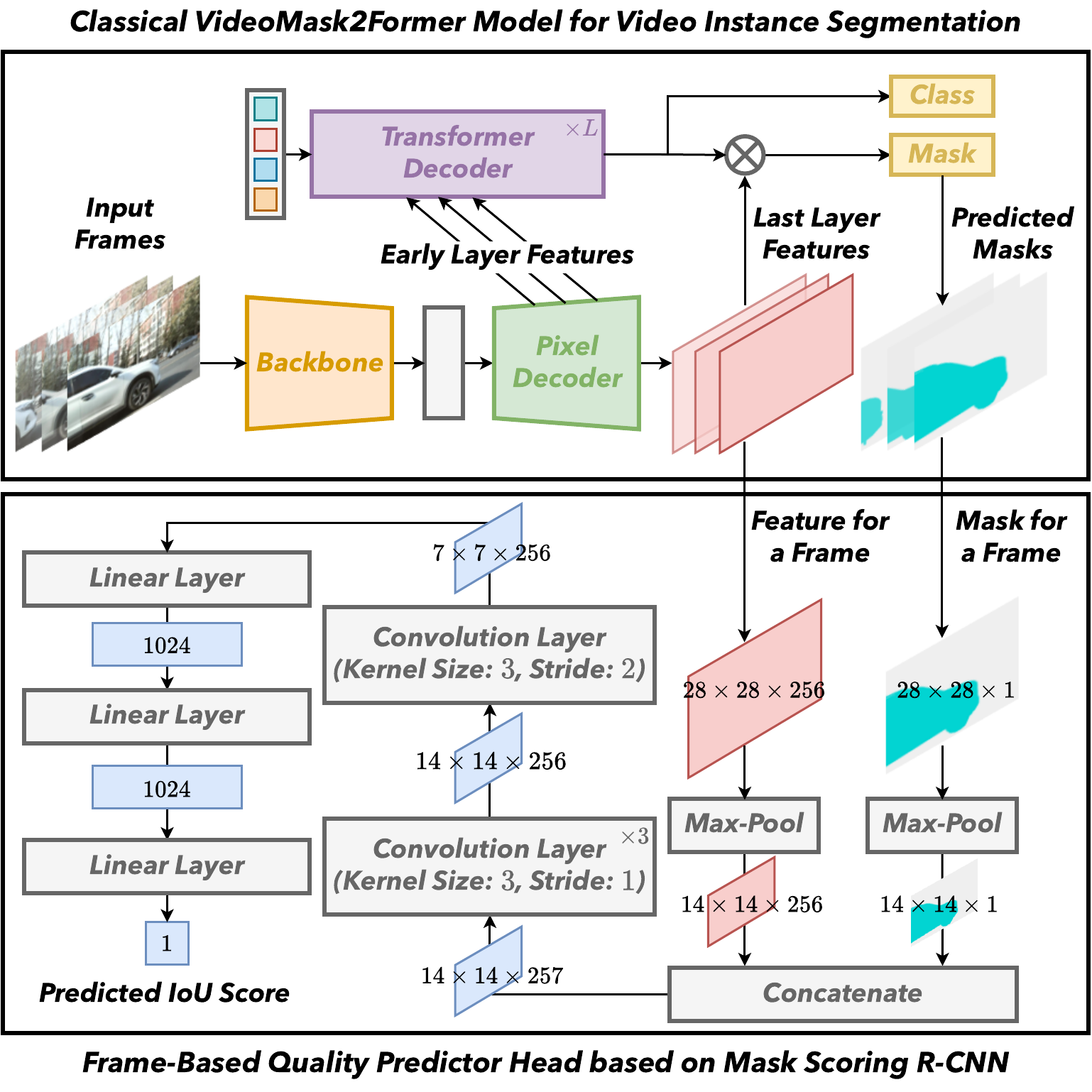}
    \caption{
    \textbf{Network architecture of VideoMask2Former~\cite{mask2former, videomask2former} and Mask Quality Predictor.} Our quality predictor integrates mask predictions and pixel decoder features following~\cite{maskscore}, employing a sequential architecture with four convolution layers (3$\times$3 kernels, final layer stride of 2 for spatial reduction) followed by three fully-connected layers that ultimately produce mask IoU predictions.}
    \label{fig:predictor}
\end{figure}

\subsection{Initial training stage}
\label{sec:initial_model}

\paragraph{Video instance segmentation (VIS) model.} Following VideoCutLER~\cite{videocutler}, we use the VideoMask2Former~\cite{mask2former, videomask2former} with the ResNet-50~\cite{resnet} backbone as our video instance segmentation (VIS) model.

\paragraph{Quality predictor.} For the quality predictor, as shown in~\cref{fig:predictor}, we use an architecture
inspired by Mask Scoring R-CNN~\cite{maskscore}. Our architecture processes individual frame features and single-object mask predictions per inference step. Supervision is established through threshold-binarized (0.5) mask IoU between predictions and matched ground truths, optimized via $\ell_2$ regression loss.

\paragraph{Synthetic videos.} VideoCutLER~\cite{videocutler} provides a high-quality pseudo-labeled synthetic video dataset that was built on the unlabeled images from ImageNet~\cite{imagenet}, which is very suitable to train and initialize our VIS model and quality predictor. We also use the trained VideoMask2Former model~\cite{videomask2former} from VideoCutLER to initialize our VIS model.

\subsection{Multi-round self-training}
\label{sec:multi_round_training}

As shown in \cref{fig:overall}, we optimize the VIS model through iterative self-training and dynamic dataset augmentation. The training dataset is initialized using the synthetic videos from VideoCutLER~\cite{videocutler}. Empirically, we find that executing model parameter restoration from the initial model weight (trained in \cref{sec:initial_model}) achieves a better performance.

After training the VIS model, we use the predicted masks on unlabeled videos with a confidence score over $0.25$ as pseudo-labels. Then we use the quality predictor to predict the IoU of each pseudo-label predicted mask. Let $\hat{\text{IoU}}_l$ denote the predicted IoU of label $l$, $s_l$ denotes the confidence score of label $l$. We define the quality score of label $l$ as $Q_l = \hat{\text{IoU}}_l \cdot s_l$.

We implement quality-based pseudo-label selection using a fixed quality score threshold $\tau_{\text{th}}$. For each pseudo-label $l$, we select it if $Q_l \geq \tau_{th}$. In the end of each round, we add all the pseudo-labels to the training dataset.


\subsection{DropLoss for mask head}
\label{sec:droploss}

\begin{table}[!t]
\resizebox{\columnwidth}{!}{%
\begin{tabular}{lrrrrrrr}
\toprule
\textbf{Method} & 
\textbf{${\text{AP}}_{50}$} &
\textbf{${\text{AP}}_{75}$} &
\textbf{${\text{AP}}$} & 
\textbf{${\text{AP}}_{S}$} & 
\textbf{${\text{AP}}_{M}$} & 
\textbf{${\text{AP}}_{L}$} & 
\textbf{${\text{AR}}_{10}$} \\ \midrule
MotionGroup~\cite{motiongroup} & 0.5  & 0.0  & 0.1  & 0.0 & 0.4  & 0.1  & 1.2  \\
OCLR~\cite{oclr}               & 5.5  & 0.3  & 1.6  & 0.1 & 1.6  & 6.1  & 11.5 \\
CutLER~\cite{cutler}           & 36.4 & 13.5 & 16.0 & 3.5 & 13.9 & 26.0 & 29.8 \\
VideoCutLER~\cite{videocutler} & 48.2 & 22.9 & 24.5 & 6.7 & 17.7 & 36.3 & 42.3 \\
AutoQ-VIS                                           & 52.6 & 28.2 & 28.1 & 6.7 & 21.2 & 40.7 & 42.5 \\
\textit{vs. previous SOTA} &
  {\color[HTML]{009901} +4.4} &
  {\color[HTML]{009901} +5.3} &
  {\color[HTML]{009901} +3.6} &
  {\color[HTML]{009901} +0.0} &
  {\color[HTML]{009901} +3.5} &
  {\color[HTML]{009901} +4.4} &
  {\color[HTML]{009901} +0.2} \\

\bottomrule
\end{tabular}%
}
\caption{
\textbf{YouTubeVIS-2019 \texttt{val}.} We reproduced MotionGroup~\cite{motiongroup}, OCLR~\cite{oclr}, CutLER~\cite{cutler}, and VideoCutLER~\cite{videocutler} results with the official code and checkpoints. AutoQ-VIS outperforms the state-of-the-art VideoCutLER by 4.4 $\text{AP}_{50}$. We evaluate results on YouTubeVIS-2019's \texttt{val} split in a class-agnostic manner.
}
\label{tab:ytvis19val}
\end{table}

We enhance the mask head training by suppressing loss contributions from low-overlap predictions, following CutLER~\cite{cutler}. For each predicted mask $m_i$, we discard its loss contribution if its maximum ground truth IoU falls below the threshold $\tau^{\text{IoU}}$:
\begin{equation}
    \mathcal{L}_{\text{drop}}(m_i) = \mathds{1}(\text{IoU}_i^{\text{max}} > \tau^{\text{IoU}})\mathcal{L}_{\text{vanilla}}(m_i)
\end{equation}

\noindent Here, $\text{IoU}_i^{\text{max}}$ is the maximum overlap between $m_i$ and any ground truth mask, while $\mathcal{L}_{\text{vanilla}}$ denotes the original mask head loss from VideoMask2Former~\cite{mask2former, videomask2former}. We employ a low threshold ($\tau^{\text{IoU}} = 0.01$) to filter only near-zero overlap predictions.

\section{Experiments}



\paragraph{Datasets.} Our model is trained on synthetic videos from VideoCutLER~\cite{videocutler} and the unlabeled \texttt{train} split of YouTubeVIS-2019~\cite{ytvis2019}. We evaluate our model's performance on the \texttt{val} split of YouTubeVIS-2019 in a class-agnostic manner.

\paragraph{Evaluation metrics.} Following~\cite{videocutler}, we use Average Precision (AP) and Average Recall (AR) as evaluation metrics. We evaluate the models in a class-agnostic manner, treating all classes as a single one during evaluation.

\paragraph{Implementation details.} For the initial training stage, we use the pretrained VideoMask2Former model~\cite{mask2former, videomask2former} from VideoCutLER~\cite{videocutler} to initialize our VIS model. Then we train the VIS model and quality predictor on synthetic videos from VideoCutLER for 8,000 iterations using a single V100 GPU, with a batch size of 2 and a learning rate of $2 \times 10^{-5}$. For each round of multi-round self-training, we train the VIS model for 10,000 iterations on two V100 GPUs, with a batch size of 4 and a learning rate of $2 \times 10^{-5}$. In practice, we find that two rounds of self-training provide the best performance.

\subsection{Experimental results}

\paragraph{Comparison with the-state-of-the-art method.} We compare AutoQ-VIS with previous top-performing methods in~\cref{tab:ytvis19val}. AutoQ-VIS achieves a remarkable improvement (about 4.4\% $\text{AP}_{50}$). Especially for $\text{AP}_{75}$, AutoQ-VIS can outperform the state-of-the-art VideoCutLER~\cite{videocutler} by 5.3\%.

\begin{table}[!t]
\resizebox{\columnwidth}{!}{%
\begin{tabular}{lrrrrrrr}
\toprule
\textbf{Method} & 
\textbf{${\text{AP}}_{50}$} &
\textbf{${\text{AP}}_{75}$} &
\textbf{${\text{AP}}$} & 
\textbf{${\text{AP}}_{S}$} & 
\textbf{${\text{AP}}_{M}$} & 
\textbf{${\text{AP}}_{L}$} & 
\textbf{${\text{AR}}_{10}$} \\ \midrule
Theoretical limit                                    & 76.8 & 48.7 & 46.8 & 13.5 & 43.6 & 62.9 & 58.0 \\
Practical limit                                & 62.7 & 33.2 & 33.9 &  4.3 & 27.3 &  53.2 & 47.5   \\
AutoQ-VIS                                           & 52.6 & 28.2 & 28.1 & 6.7  & 21.2 & 40.7 & 42.5 \\ 
\bottomrule
\end{tabular}%
}
\caption{\textbf{Comparison with the theoretical and practical limit.} 
\textit{Theoretical Limit}: Upper-bound performance achieved by training on ground-truth labels from YouTubeVIS-2019 \texttt{train} split in class-agnostic mode, representing ideal supervision conditions.  
\textit{Practical Limit}: Best attainable performance when using all pseudo-labels with $\text{IoU}\geq0.5$ against ground truth, simulating perfect pseudo-label selection.
}
\label{tab:limit_comparison}
\end{table}

\paragraph{Comparison with the theoretical and practical limit.} \cref{tab:limit_comparison} reveals a significant performance gap (10.1 $\text{AP}_{50}$) between AutoQ-VIS and the practical upper bound, indicating substantial potential for improvement through enhanced pseudo-label utilization. The theoretical limit represents a fully supervised training using all ground-truth annotations from YouTubeVIS-2019
\texttt{train} set. The practical limit represents an oracle experiment that simulates perfect pseudo-label selection by using all predictions with $\text{IoU} \geq 0.5$ against ground truth.


\begin{figure}[t]
    \centering
    \includegraphics[width=1\linewidth]{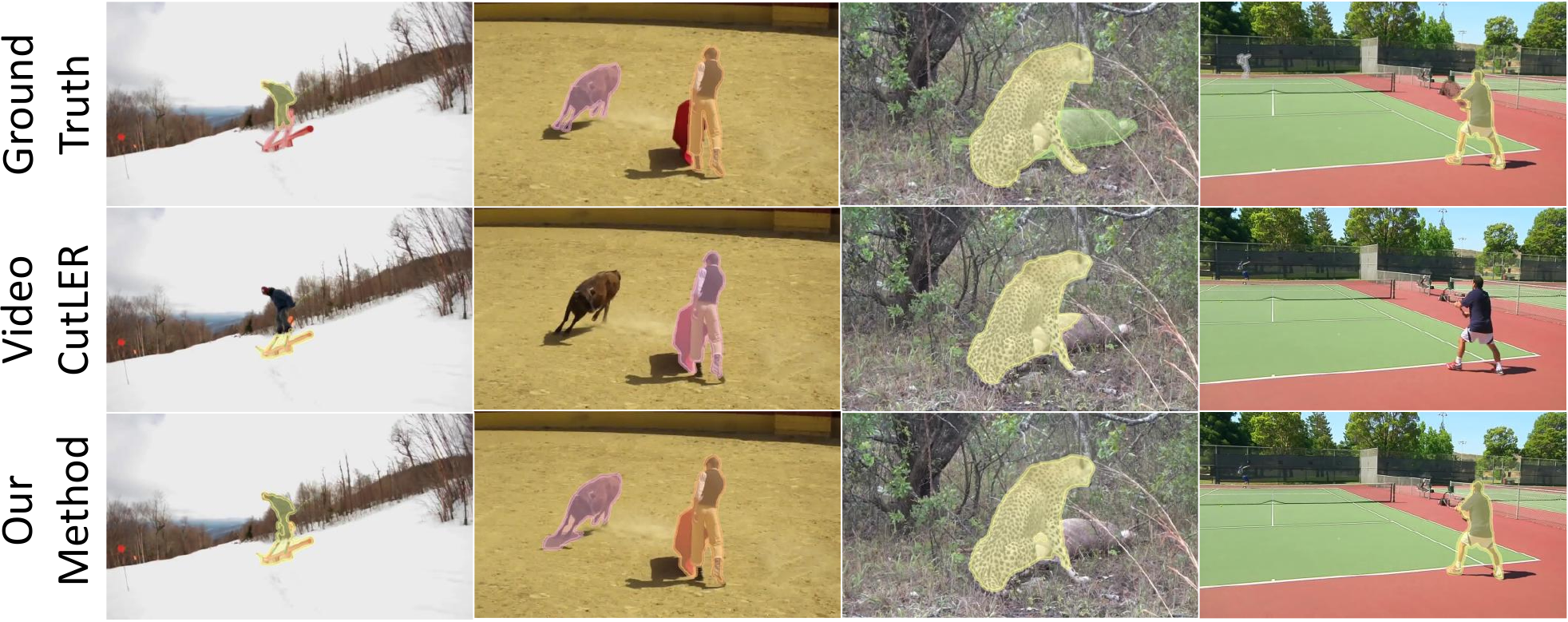}
    \caption{\textbf{The qualitative results on YouTubeVIS-2019 \texttt{val} split.} AutoQ-VIS demonstrates superior instance discovery capabilities compared to VideoCutLER~\cite{videocutler}: (1) Enhanced multi-object detection capacity, particularly for semantically distinct instances (e.g., person and bull in Column 2); (2) Improved segmentation fidelity through precise boundary delineation (e.g., the leopard in Column 3). (3) Better comprehensive instance coverage, eliminating false negatives (e.g., detecting humans in Columns 1 \& 4 that VideoCutLER completely misses, even without occlusion or scale challenges).
    }
    \label{fig:qualitative_results}
\end{figure}

\paragraph{Qualitative results.} Fig.~\ref{fig:qualitative_results} demonstrates AutoQ-VIS's advancements over VideoCutLER~\cite{videocutler}. As we observe, AutoQ-VIS is capable of discovering more objects and producing higher-quality segmentation masks.

\subsection{Ablation studies}

\begin{table}[t]
\resizebox{\columnwidth}{!}{%
\begin{tabular}{lrrrrrrr}
\toprule
\textbf{Method} & 
\textbf{${\text{AP}}_{50}$} &
\textbf{${\text{AP}}_{75}$} &
\textbf{${\text{AP}}$} & 
\textbf{${\text{AP}}_{S}$} & 
\textbf{${\text{AP}}_{M}$} & 
\textbf{${\text{AP}}_{L}$} & 
\textbf{${\text{AR}}_{10}$} \\ \midrule
w/o quality predictor                            & 50.5 & 25.9 & 27.2 & 5.7 & 19.0 & 40.5 & \textbf{43.3} \\
w/o DropLoss                                     & 48.0 & 23.7 & 24.6 & 3.8 & 16.9 & 37.8 & 39.2 \\
w/o resetting each round                        & 51.6 & 28.0 & \textbf{28.2} & 6.4 & \textbf{22.8} & 40.6 & 42.9 \\
AutoQ-VIS                                           & \textbf{52.6} & \textbf{28.2} & 28.1 & \textbf{6.7} & 21.2 & \textbf{40.7} & 42.5 \\ 
\bottomrule
\end{tabular}%
}
\caption{
\textbf{Ablation study on the contribution of each component.} \textit{Without quality predictor:} We remove the quality predictor, and use the confidence score $s_l$ as quality score $Q_l$ with threshold $\tau_{th} = 0.85$. \textit{Without DropLoss:} We use the vanilla loss for the mask head instead of the DropLoss. \textit{Without resetting each round:} The model weights are not reset at the beginning of each round.
}
\label{tab:component_ablation}
\end{table}

\paragraph{Component ablation analysis.} \cref{tab:component_ablation} quantifies individual component contributions through progressive additions. The DropLoss provides the most substantial gains (+4.6 $\text{AP}_{50}$), followed by the quality predictor's +2.1 $\text{AP}_{50}$ improvement. Notably, even the confidence score baseline surpasses VideoCutLER by +2.3 $\text{AP}_{50}$, demonstrating fundamental advantages of our self-training method. While model resetting yields marginal gains (+1.0 $\text{AP}_{50}$, +0.2 $\text{AP}_{75}$), it maintains baseline $\text{AP}$ performance.

\begin{table}[t]
\resizebox{\columnwidth}{!}{%
\begin{tabular}{lrrrrrrr}
\toprule
\textbf{Self-training} & 
\textbf{${\text{AP}}_{50}$} &
\textbf{${\text{AP}}_{75}$} &
\textbf{${\text{AP}}$} & 
\textbf{${\text{AP}}_{S}$} & 
\textbf{${\text{AP}}_{M}$} & 
\textbf{${\text{AP}}_{L}$} & 
\textbf{${\text{AR}}_{10}$} \\ \midrule
1 round                                                                      & 51.3 & 26.5 & 26.9 & \textbf{7.0} & \textbf{22.3} & 38.4 & \textbf{43.2} \\
2 rounds                                                                      & \textbf{52.6} & \textbf{28.2} & \textbf{28.1} & 6.7 & 21.2 & \textbf{40.7} & 42.5 \\
3 rounds                                                                      & 52.0 & 27.0 & 27.7 & 6.2 & 21.8 & 40.1 & 42.1 \\
\bottomrule
\end{tabular}%
}
\caption{
\textbf{Ablation study on the number of self-training rounds.} Our framework achieves peak performance (52.6 $\text{AP}_{50}$) at the second round before gradual degradation (-0.6 $\text{AP}_{50}$) from pseudo-label noise accumulation. This establishes round 2 as the optimal stopping point to balance accuracy and error propagation risks.
}
\label{tab:round_num}
\end{table}

\paragraph{Self-training round analysis.} \cref{tab:round_num} tests how many self-training rounds work best. The model hits its peak (52.6 $\text{AP}_{50}$) at round 2, then slowly gets worse. This drop occurs because, as we perform more rounds, mistakes in the pseudo-labels pile up.

\begin{table}[t]
\resizebox{\columnwidth}{!}{%
\begin{tabular}{rrrrrrrr}
\toprule
\textbf{$\tau_{th}$} & 
\textbf{${\text{AP}}_{50}$} &
\textbf{${\text{AP}}_{75}$} &
\textbf{${\text{AP}}$} & 
\textbf{${\text{AP}}_{S}$} & 
\textbf{${\text{AP}}_{M}$} & 
\textbf{${\text{AP}}_{L}$} & 
\textbf{${\text{AR}}_{10}$} \\ \midrule
0.95 & 48.7 & 25.3 & 26.1 & 5.9 & 18.8 & 38.5 & 40.7 \\
0.85 & 48.8 & 24.6 & 26.0 & 5.8 & 18.6 & 38.6 & 41.2 \\
0.75 & \textbf{52.6} & \textbf{28.2} & \textbf{28.1} & \textbf{6.7} & \textbf{21.2} & \textbf{40.7} & \textbf{42.5} \\
0.50  & 52.4 & 25.7 & 27.1 & 6.5 & 20.1 & 39.9 & 42.2 \\ \bottomrule
\end{tabular}%
}
\caption{
\textbf{Ablation study on the quality score threshold $\tau_{th}$.} Optimal performance (52.6 $\text{AP}_{50}$) emerges at $\tau_{th}=0.75$, balancing valid sample retention and noise suppression. Lower threshold ($\tau_{th}=0.50$) degrades results by admitting too many low-quality predictions, while the higher thresholds ($\tau_{th}=0.95$ and $\tau_{th}=0.80$) oversuppress valid samples.
}
\label{tab:quality_threshold}
\end{table}

\paragraph{Quality score threshold $\tau_{th}$ analysis.} \cref{tab:quality_threshold} examines the impact of quality score thresholds on pseudo-label selection. We observe a non-monotonic relationship: While lower thresholds ($\tau_{th} \leq 0.75$) generally yield superior performance by retaining more valid samples, excessively lenient selection ($\tau_{th}=0.50$) introduces noisy supervision, degrading results. The optimal balance occurs at $\tau_{th}=0.75$, achieving peak performance.



\begin{figure}[t]
    \centering
    \begin{subfigure}[b]{0.49\linewidth}
        \includegraphics[width=\linewidth]{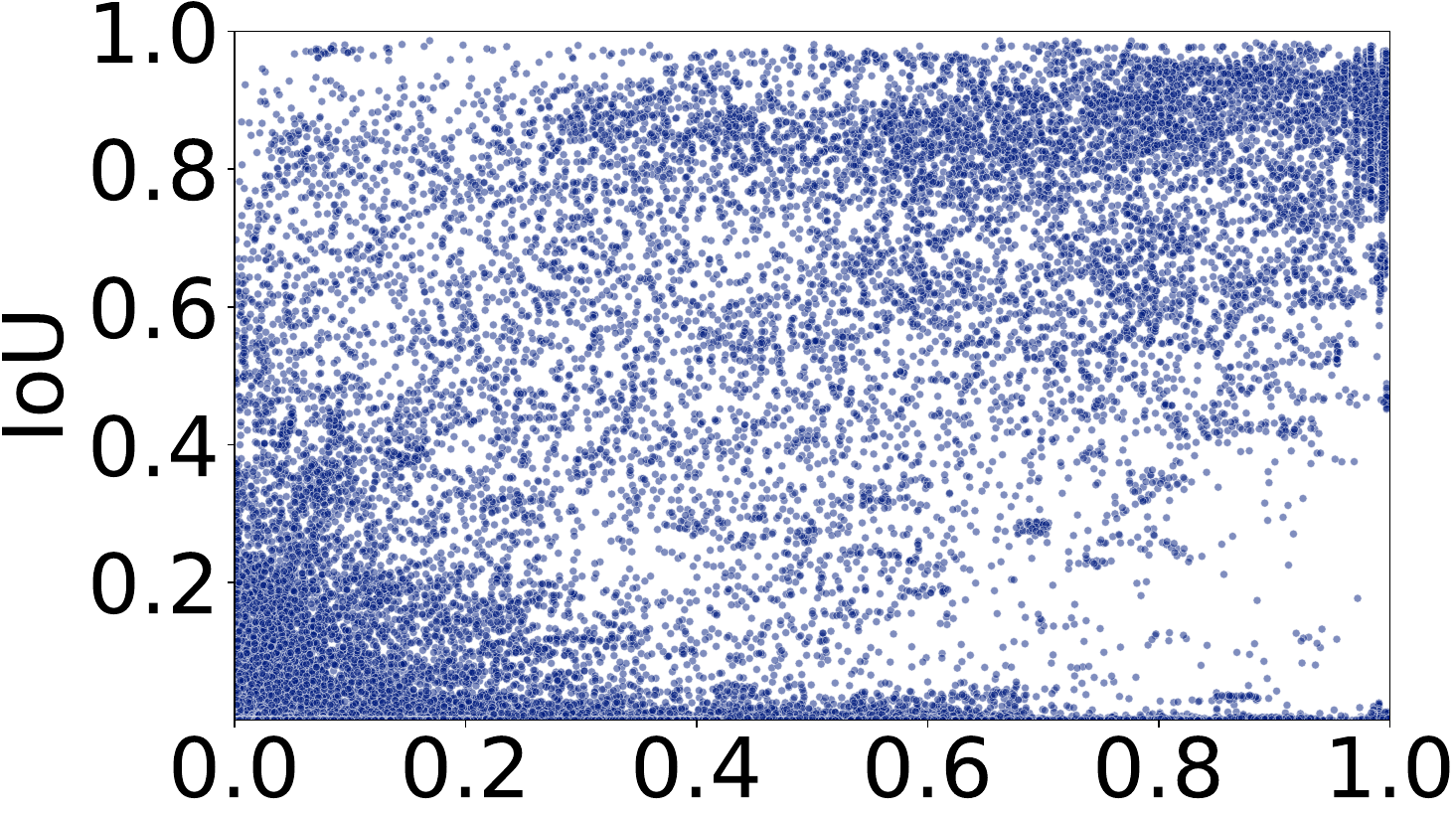}
        \caption*{\parbox[c]{\linewidth}{\centering Quality score, $Q_l$ ($\rho_{s}=0.57$)}}
    \end{subfigure}
    \hfill
    \begin{subfigure}[b]{0.49\linewidth}
        \includegraphics[width=\linewidth]{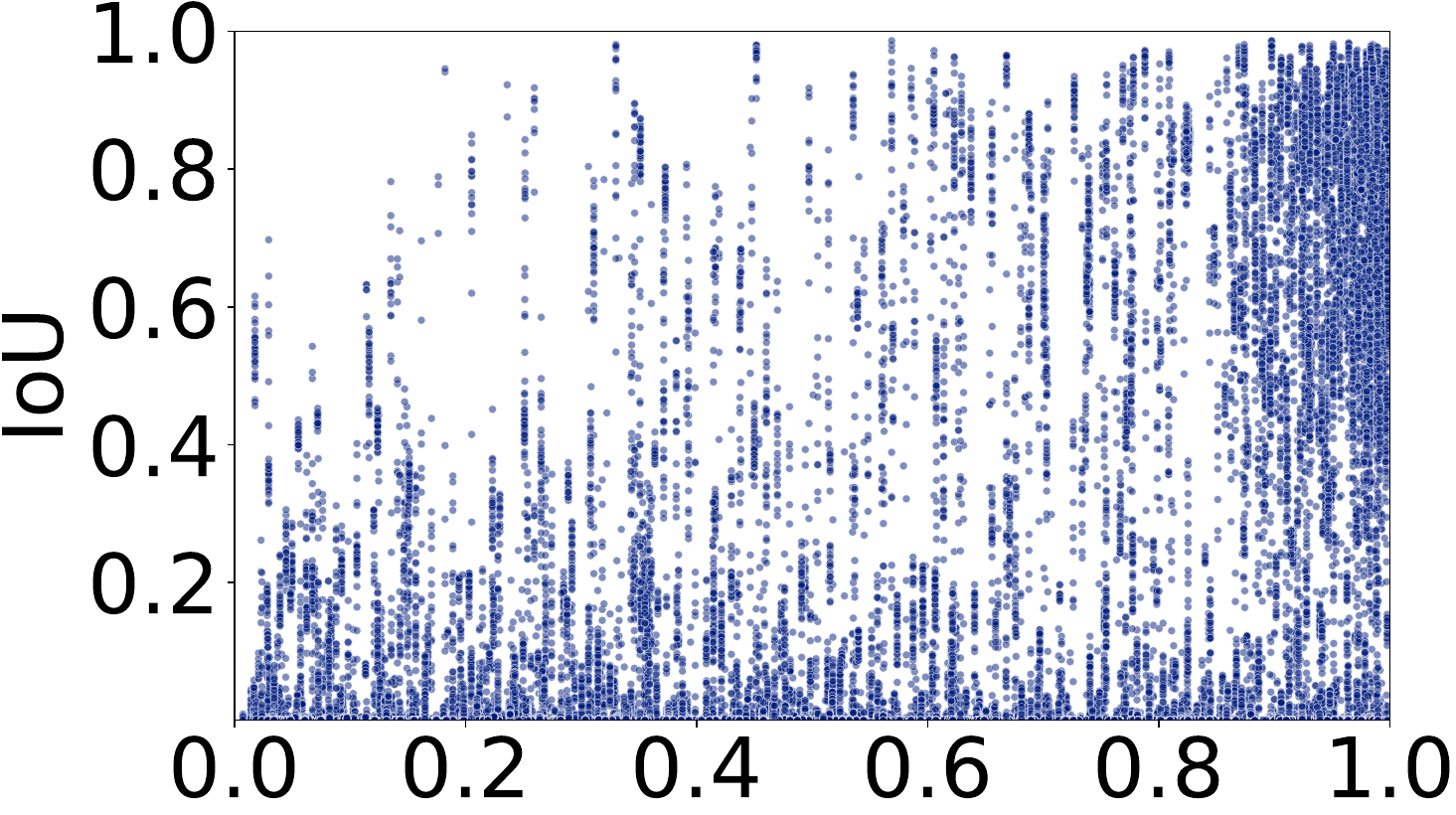}
        \caption*{\parbox[c]{\linewidth}{\centering Confidence score, $s_l$ ($\rho_{s}=0.42$)}}
    \end{subfigure}
    \caption{
    \textbf{Visualized comparison of quality score $Q_l$ and confidence score $s_l$.} Here, $\rho_{s}$ denotes the Spearman's rank correlation coefficient. Subplot (a) visualizes quality scores $Q_l$ and their ground truth IoU. Subplot (b) visualizes confidence scores $s_l$ and their ground truth IoU.}
    \label{fig:distribution}
\end{figure}


\paragraph{Quality score vs. confidence score.} As shown in \cref{fig:distribution},
our quality score $Q_s$ has a higher correlation to the IoU of the pseudo label and the ground truth label than the confidence score of VideoMask2Former, which proves our quality predictor's effectiveness in pseudo-label quality assessment. This results in a significant improvement in the final VIS model performance (+2.1 $\text{AP}_{50}$) in \cref{tab:component_ablation}.
\section{Conclusion}

We present AutoQ-VIS, a quality-aware self-training framework that advances unsupervised video instance segmentation through iterative pseudo-label refinement with automatic quality control. By establishing a closed-loop system of pseudo-label generation and automatic quality assessment, our method achieves state-of-the-art performance (52.6 $\text{AP}_{50}$) on YouTubeVIS-2019 \texttt{val} split without requiring any human annotations. The simple quality predictor proves effective in pseudo-label quality assessment. 

\onur{I did not like the way you formulate the main contribution: "bridging the synthetic-to-real domain gap". This sounds like a domain adaptation paper but our method is about sth. else. Can we rephrase it?}


\onur{refine the references, inconsistent formating}
\clearpage\newpage
{
    \small
    \bibliographystyle{ieeenat_fullname}
    \bibliography{main}
}

\clearpage
\setcounter{page}{1}
\maketitlesupplementary


\section{Detailed methodology}
\label{detailed_method}

This section supplements what is not clearly stated in \cref{sec:multi_round_training}.

\subsection{Automated pseudo-annotation with spatio-temporal NMS}

After the training of the VIS model, we use it to label the unlabeled videos. 
Let $\mathcal{D} = \{d_i\}^{N}_{i=1}$ denote the initial detection set per video, where each detection $d = (s_i, \{m_i^t\}^{T}_{t=1})$ contains:
\begin{itemize}
    \item $s_i \in [0,1]$: Confidence score
    \item $\{m_i^t\}^{T}_{t=1}$: Binary mask sequence across $T$ frames
\end{itemize}

We filter those detections that have confidence scores larger than or equal to 0.25:
\begin{equation}
    \mathcal{D}_{\text{filtered}} = \{ d_i \in \mathcal{D} \mid s_i \geq 0.25 \}
\end{equation}

However, the detection sets may contain duplicate detections. To solve this problem, we need to perform spatiotemporal non-maximum suppression. First, we sort the detections based on their confidence scores:
\begin{equation}
    \mathcal{D}_{\text{sorted}} = \{ d_{(k)} \}_{k=1}^{|\mathcal{D}_{\text{filtered}}|} \quad \text{s.t.} \quad \forall i < j: s_{(i)} \geq s_{(j)}
\end{equation}

Our method eliminates redundant detections through spatiotemporal overlap analysis: Any lower-confidence prediction is suppressed if exhibiting mask overlap (IoU $\geq$ 0.5) with higher-confidence detections in at least one video frame. Formally, let $\mathcal{D}_{\text{suppressed}} \subseteq \mathcal{D}_{\text{sorted}}$ represent the preserved detection set after suppression:

\begin{equation}
\begin{split}
d_{(k)} \in \mathcal{D}_{\text{suppressed}} \iff & \nexists d_{(p)} \in \mathcal{D}_{\text{suppressed}} \ \text{where} \ p < k \ \text{s.t.} \\
& \exists t \in [1,T] : \frac{\|m_{(k)}^t \cap m_{(p)}^t\|}{\|m_{(k)}^t \cup m_{(p)}^t\|} \geq 0.5 \\
& \underbrace{\hphantom{\exists t \in [1,T] : \frac{\|m_{(k)}^t \cap m_{(p)}^t\|}{\|m_{(k)}^t \cup m_{(p)}^t\|}}}_{\mathclap{\text{Frame-specific overlap condition}}}
\end{split}
\end{equation}
where $\| \cdot \|$ denotes pixel cardinality. This temporal-existential criterion suppresses duplicates appearing in any frame of the video sequence.

\subsection{Confidence-aware filtration via quality predictor}

Let $\mathcal{D}_{\text{global}}^{(k)}$ denote the union of $\mathcal{D}_{\text{suppressed}}$ of all videos:
\begin{equation}
    \mathcal{D}_{\text{global}}^{(k)} = \bigcup_{v \in \mathcal{V}} \mathcal{D}_{\text{suppressed},v}^{(k)}
\end{equation}
where $\mathcal{D}_{\text{suppressed},v}^{(k)}$ denotes preserved detections for video $v$ after spatiotemporal NMS at iteration $k$.

For each detection $d \in \mathcal{D}_{\text{global}}^{(k)}$, we define:
\begin{equation}
    Q_d^t = s_d \cdot \hat{\text{IoU}}_d^t \quad \forall t \in [1,T_v]
\end{equation}
where $Q_d^t$ is the quality score of detection $d$ in frame $t$, $s_d$ is the confidence score of detection $d$, and $T_v$ denotes the number of frames in video $v$.

We implement quality-based pseudo-label selection using a fixed quality threshold $\tau_{\text{th}}$. For each detection $d \in \mathcal{D}_{\text{global}}^{(k)}$ across all videos:

\begin{equation}
    \mathcal{S}_d^t = \begin{cases} 
1 & Q_d^t \geq \tau^{(k)} \\
0 & \text{otherwise}
\end{cases}
\end{equation}
where $\mathcal{S}_d^t$ denote whether pseudo-label of detection $d$ in frame $t$ is selected. For each detection, if its results are not selected in any frames, we discard it:

\begin{equation}
\mathcal{D}_{\text{retained}}^{(k)} = \left\{ d_v \in \mathcal{D}_{\text{global}}^{(k)} \,\bigg|\, \sum_{t=1}^{T_v} \mathcal{S}_d^t > 0 \right\}
\end{equation}
where $\mathcal{D}_{\text{retained}}^{(k)}$ is the set of detections we retain in iteration $k$.

\subsection{Dynamic dataset augmentation through adaptive fusion of newly curated annotations}

The augmentation process operates in two distinct modes depending on whether the video containing the new detections already exists in the training set. For novel videos, we perform bulk insertion of all qualified detections. For existing videos, we implement an instance-level fusion protocol that intelligently merges new predictions with historical annotations while preserving temporal consistency.

Let $\mathcal{T}^{(k)} = \{(v, \mathcal{D}_v)\}$ denote the training dataset at iteration $k$, where $\mathcal{D}_v$ contains all preserved detections for video $v$. We need to merge the $\mathcal{D}_{\text{retained}}^{(k)}$ into the $\mathcal{T}^{(k)} = \{(v, \mathcal{D}_v)\}$. For each video, if it's not in the training dataset, we simply add all its detections to the training dataset. Formally, let $\mathcal{D}_{\text{retained},v}^{(k)}$ denote the detections belong to video $v$:
\begin{equation}
    \mathcal{D}_{\text{retained},v}^{(k)} = \left\{ d \in \mathcal{D}_{\text{retained}}^{(k)} \,\big|\, d \text{ belongs to video } v \right\}
\end{equation}

Let $\mathcal{V}_{\text{new}}^{(k)} = \pi_1(\mathcal{D}_{\text{retained}}^{(k)}) \setminus \pi_1(\mathcal{T}^{(k)})$ denote completely new videos, where $\pi_1$ projects to video identifiers. Let $\mathcal{T}_{\text{new}}^{(k)}$ denote labels of new videos:

\begin{equation}
\mathcal{T}_{\text{new}}^{(k+1)} =  \bigcup_{v \in \mathcal{V}_{\text{new}}^{(k)}} \left\{ \left(v, \mathcal{D}_{\text{retained},v}^{(k)} \right) \right\}
\end{equation}

For videos already present in the training set, we implement a temporal-aware fusion protocol that resolves conflicts between new detections and historical annotations.

To merge the existing detections and new detections, we need to identify whether two detections overlap. We define the spatiotemporal overlap predicate for two detections:
\begin{equation}
    \phi(d_{\text{new}, v}, d_{\text{exist}, v}) \triangleq \exists t \in [1, T_v]: \frac{\|m_{\text{new}}^t \cap m_{\text{exist}}^t\|}{\|m_{\text{new}}^t \cup m_{\text{exist}}^t\|} \geq 0.5
\end{equation}

This predicate is very similar to what we use in spatiotemporal NMS. If in any frames, two detections' masks overlap (IoU $\geq$ 0.5), we consider them overlapped detections. If two detections overlap, we need to fuse them. We fuse the new detection and the existing detection frame by frame. For one frame, if only the existing detection's label is selected, we use its label; otherwise, we use the new detection's label. Let $\mathcal{F}$ denote the fusion operation:
\begin{equation}
    \mathcal{F}(d_{\text{new},v}, d_{\text{exist},v}) =  \{\mathcal{S}_{\text{merge}}^t, m_{\text{merge}}^t\}_{t=1}^{T_v}
\end{equation}
where:
\begin{equation}
    \mathcal{S}_{\text{merge}}^t = \max(\mathcal{S}_{\text{new}}^t, \mathcal{S}_{\text{exist}}^t)
\end{equation}
\begin{equation}
    \forall t \leq T_{\text{merge}}: m_{\text{merge}}^t = \begin{cases} 
m_{\text{exist}}^t & \text{if } \mathcal{S}_{\text{exist}}^t = 1 \land \mathcal{S}_{\text{new}}^t = 0 \\
m_{\text{new}}^t & \text{otherwise}
\end{cases}
\end{equation}

If the new detection doesn't overlap with any existing detections, we simply add it. Let $\mathcal{V}_{\text{exist}}^{(k)} = \pi_1(\mathcal{T}^{(k)})$ denote existing videos. For each existing video $v \in \mathcal{V}_{\text{exist}}^{(k)}$, we process all new detections $\mathcal{D}_{\text{retained},v}^{(k)}$ through:
\begin{equation}
    \mathcal{D}_v^{(k+1)} = \mathcal{D}_v^{(k)} \bigoplus \mathcal{D}_{\text{retained},v}^{(k)}
\end{equation}
where the fusion operation $\bigoplus$ is defined as:
\begin{equation}
    \mathcal{D} \bigoplus \mathcal{D}' = \bigcup_{d' \in \mathcal{D}'} \left( \mathcal{D} \oplus d' \right)
\end{equation}
with per-detection fusion:

\begin{equation}
\mathcal{D} \oplus d_{\text{new}} = 
\begin{cases}
\mathcal{D} \cup \{d_{\text{new}}\} & \text{if } \forall d_{\text{exist}} \in \mathcal{D}, \\
& \quad \neg\phi(d_{\text{new}}, d_{\text{exist}}) \\
\mathcal{D} \setminus d_{\text{exist}} \cup \mathcal{F}(d_{\text{new}}, d_{\text{exist}}) & \text{if } \exists d_{\text{exist}} \in \mathcal{D}, \\
& \quad \phi(d_{\text{new}}, d_{\text{exist}})
\end{cases}
\end{equation}

In the end, we merge the labels of new videos and existing videos:
\begin{equation}
    \mathcal{T}^{(k+1)} = \mathcal{T}_{\text{new}}^{(k+1)} \cup \mathcal{T}_{\text{exist}}^{(k+1)}
\end{equation}
where:
\begin{equation}
    \mathcal{T}_{\text{exist}}^{(k+1)} =  \bigcup_{v \in \mathcal{V}_{\text{exist}}^{(k)}} \left\{ \left(v, \mathcal{D}_{v}^{(k+1)} \right) \right\}
\end{equation}

During the training, for each video, we need to sample 3 frames as the input of the model. To provide good-quality labels, we only sample those frames where all their detections are selected. For each video $v \in \mathcal{V}_{\text{train}}^{(k+1)}$, we define the eligible frame set:
\begin{equation}
\mathcal{F}_{\text{eligible}}^{(k+1)}(v) = \left\{ t \in [1,T_v] \,\big|\, \forall d \in \mathcal{D}_v^{(k+1)}: \mathcal{S}_d^t = 1 \right\}
\end{equation}

The training batch is constructed through uniform sampling:
\begin{equation}
\mathcal{B}_{\text{train}}^{(k+1)}(v) = \mathop{\mathrm{UniformSample}}\limits_{3} \left( \mathcal{F}_{\text{eligible}}^{(k+1)}(v) \right)
\end{equation}

Although we only use a subset of pseudo-labels, all pseudo-labels persist in the training set regardless of selection status, enabling progressive refinement.

To balance data distribution between VideoCutLER's~\cite{videocutler} extensive synthetic videos and our pseudo-labeled videos, we implement a balanced stochastic sampling strategy. Each training batch has equal probability (50\%) of being drawn from either the synthetic videos or the pseudo-labeled set. This prevents overfitting to either domain while maintaining the synthetic data's regularization benefits during self-training.

\section{Additional qualitative visualizations}

\begin{figure*}[h]
    \centering
    \includegraphics[width=1\linewidth]{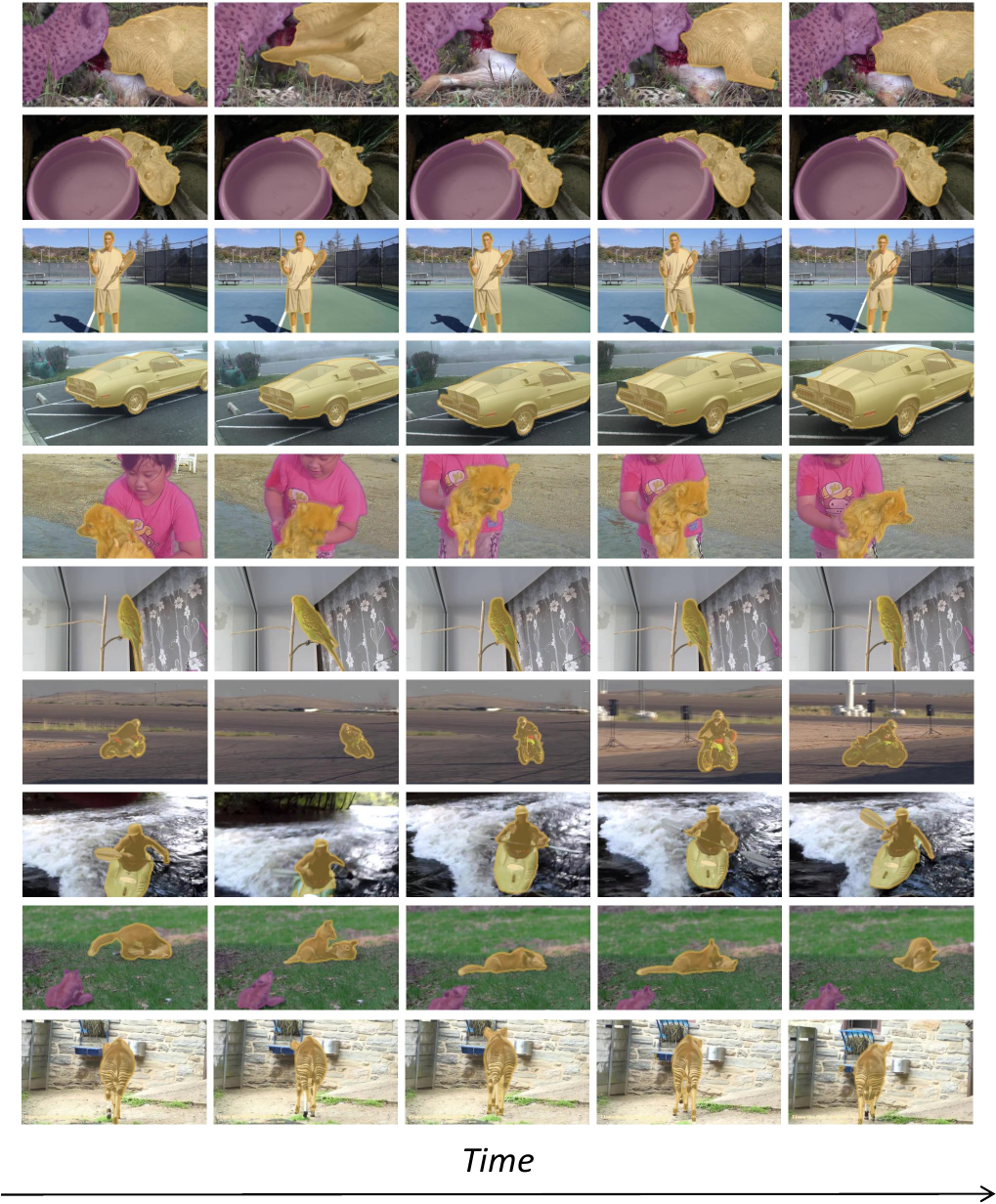}
    \caption{\textbf{Qualitative results of our VIS model on YouTubeVIS-2019 \texttt{val} split.}}
    \label{fig:add_vis_visual}
\end{figure*}

\begin{figure*}[h]
    \centering
    \includegraphics[width=1\linewidth]{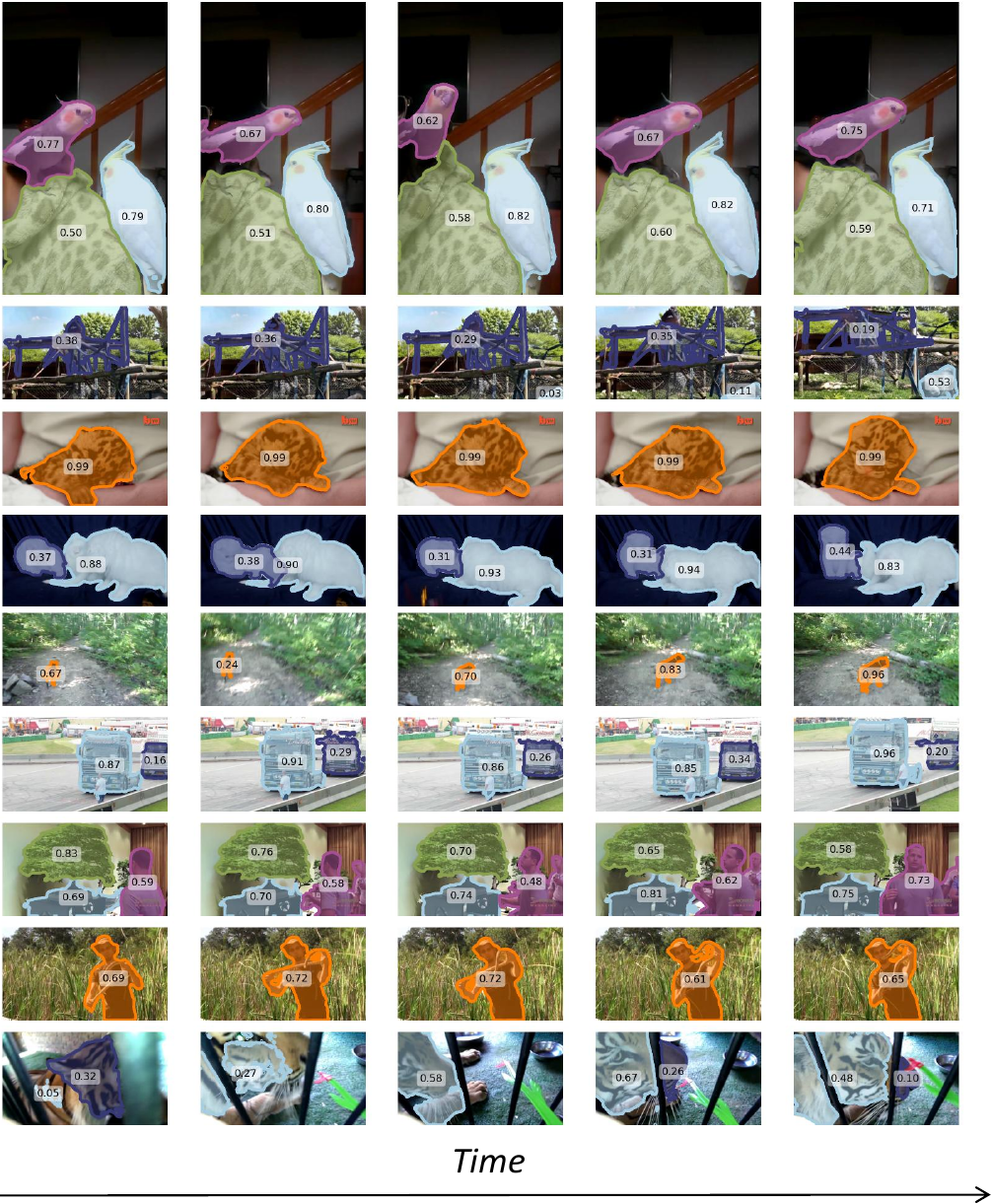}
    \caption{\textbf{Qualitative results of our quality predictor on YouTubeVIS-2019 \texttt{train} split.} The quality scores are shown in the center of each pseudo label.}
    \label{fig:quality_visual}
\end{figure*}

We provide additional qualitative results of our VIS model and quality predictor in \cref{fig:add_vis_visual} and \cref{fig:quality_visual}.

\end{document}